\documentclass[conference]{IEEEtran}
\IEEEoverridecommandlockouts

\usepackage{cite}
\usepackage{amsmath,amssymb,amsfonts}
\usepackage{algorithmic}
\usepackage{graphicx}
\usepackage{textcomp}
\usepackage{xcolor}
\usepackage{caption}
\usepackage{algorithm}
\usepackage{threeparttable}
\usepackage{multicol}
\usepackage{multirow}
\usepackage{booktabs}
\def\BibTeX{{\rm B\kern-.05em{\sc i\kern-.025em b}\kern-.08em
    T\kern-.1667em\lower.7ex\hbox{E}\kern-.125emX}}
\begin{document}

\title{Revisiting the Privacy Risks of Split Inference: A GAN-Based Data Reconstruction Attack via Progressive Feature Optimization}
\renewcommand\footnotemark{}
\author{
Yixiang Qiu$^{2}$, Yanhan Liu$^{2}$, Hongyao Yu$^{2}$, Hao Fang$^{2}$, Bin Chen$^{1\ddagger}$, Shu-Tao Xia$^{2}$, Ke Xu$^{3}$ \\
$^{1}$School of Computer Science and Technology, Harbin Institute of Technology, Shenzhen \\
$^{2}$Tsinghua Shenzhen International Graduate School, Tsinghua University   \\
$^{3}$Department of Computer Science and Technology, Tsinghua University \\
$^{\ddagger}$Corresponding author: Bin Chen (chenbin2021@hit.edu.cn) \\
\footnotesize{\texttt{\{qiu-yx24,liuyanha25,hy-yu25,fang-h23\}@mails.tsinghua.edu.cn;}} \\
\footnotesize{\texttt{chenbin2021@hit.edu.cn;}}
\footnotesize{\texttt{xiast@sz.tsinghua.edu.cn;}}
\footnotesize{\texttt{xuke@tsinghua.edu.cn}}
}

\maketitle

\newcommand{\todo}[1]{\textcolor{blue}{TODO: #1.}}
\begin{abstract}
The growing complexity of Deep Neural Networks (DNNs) has led to the adoption of Split Inference (SI), a collaborative paradigm that partitions computation between edge devices and the cloud to reduce latency and protect user privacy. However, recent advances in Data Reconstruction Attacks (DRAs) reveal that intermediate features exchanged in SI can be exploited to recover sensitive input data, posing significant privacy risks. Existing DRAs are typically effective only on shallow models and fail to fully leverage semantic priors, limiting their reconstruction quality and generalizability across datasets and model architectures. In this paper, we propose a novel GAN-based DRA framework with \textit{Progressive Feature Optimization (PFO)}, which decomposes the generator into hierarchical blocks and incrementally refines intermediate representations to enhance the semantic fidelity of reconstructed images. To stabilize the optimization and improve image realism, we introduce an $\ell_1$-ball constraint during reconstruction. Extensive experiments show that our method outperforms prior attacks by a large margin, especially in high-resolution scenarios, out-of-distribution settings, and against deeper and more complex DNNs.
\end{abstract}
\section{Introduction}
In recent years, Deep Neural Networks (DNNs) have undergone rapid development and achieved tremendous success in various applications, such as face recognition \cite{he2016deep}, personalized recommendation \cite{wu2017session}, and privacy computing \cite{yu2022thwarting}. However, alongside the continuous enhancement in model performance, the growing size of model parameters has become a burden for edge devices. As a result, it is challenging to deploy these models on resource-constrained edge devices while satisfying efficiency requirements (e.g., computational cost and response delay). Meanwhile, users utilizing cloud APIs raise significant concerns about privacy leakage, as raw data is collected by cloud service providers for service provisioning.

In this context, Split Inference (SI) \cite{hauswald2014hybrid,eshratifar2019jointdnn,banitalebi2021auto,kang2017neurosurgeon,matsubara2022split,ko2024two} has emerged as a promising solution since it balances efficiency with privacy. Specifically, SI enables model providers to partition a large-scale DNN into two segments, deploying them on edge devices and cloud servers, respectively, to facilitate collaborative inference between the cloud and edge devices \cite{ko2024two}. Each user’s private raw data undergoes preprocessing on the partial model at the edge device before the corresponding intermediate representations are transmitted to the cloud server for subsequent inference. Therefore, SI prevents cloud service providers from accessing raw data directly, ensuring the privacy and security of users. Moreover, as both the edge devices and cloud server execute only a portion of the full model, SI alleviates resource constraints on edge devices while fully leveraging the abundant storage and computational resources of the cloud server.

\begin{figure}[t]
\centerline{\includegraphics[width=\columnwidth]{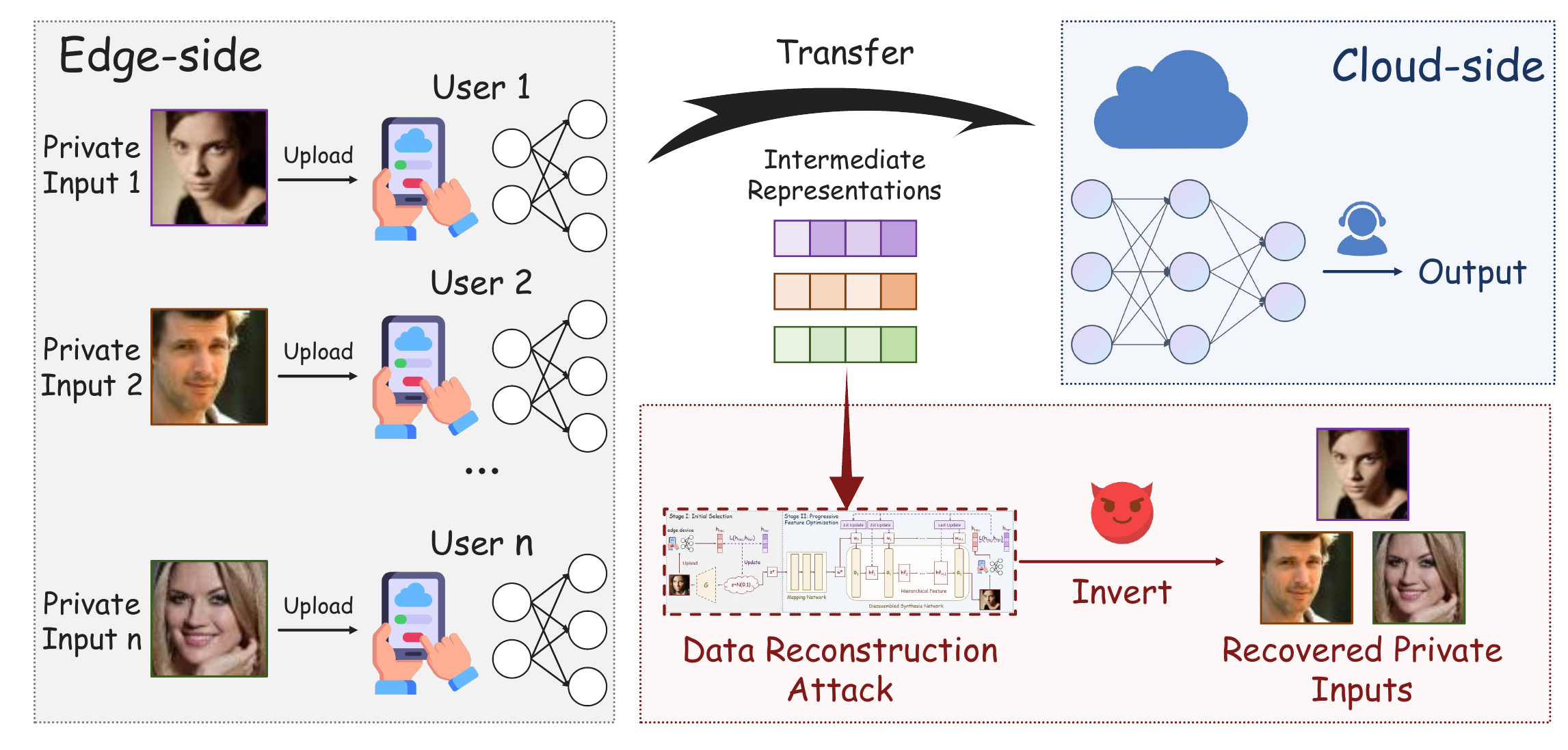}}
\caption{An overview of data reconstruction attacks on the split inference system. The malicious attacker intercepts the intermediate representations and tries to invert private facial information via pixel-level reconstruction of target images.}
\label{fig:dra}
\end{figure}

Despite the aforementioned advantages, recent studies \cite{dong2021privacy,sa2024ensuring,fang2024privacy,qiu2024mibench1,qiu2024mibench2,qiu2024mibench3,yu2025gi,yu2024editable,chen2024editable,fang2024one} have revealed that SI systems still pose threats of privacy leakage. Among these attack methods, the Data Reconstruction Attack (DRA) \cite{he2019model,singh2021disco,yang2022measuring,li2023gan,xu2024stealthy} is known as a representative threat due to its powerful capability in recovering the privacy-sensitive data. As shown in Fig. \ref{fig:dra}, malicious attackers (e.g., a curious service provider who desires to obtain the privacy from users) aim to exploit the intermediate representations exchanged between the cloud and edge devices to reconstruct the user's input data, thereby recovering the targeted privacy-sensitive information. Given that DRAs can be launched throughout the entire model release lifecycle and achieve pixel-level image reconstruction, they tend to pose a more threatening nature to users' privacy.

However, existing DRA methods suffer from several limitations that significantly diminish their effectiveness and practicality. For instance, current DRAs \cite{he2019model,singh2021disco,yang2022measuring,li2023gan,xu2024stealthy} can only attack shallow models with simple architectures (e.g., LeNet \cite{lecun2002gradient}) and reconstruct images with relatively low resolution (i.e., $64\times64$ pixels), falling short of adapting to more complex model architectures and higher-resolution images encountered in real-world scenarios. Moreover, these DRAs exhibit weak support for Out-of-Distribution (OOD) scenarios, where there is a significant distributional shift between the target private dataset and the public dataset utilized as auxiliary prior \cite{he2019model}. 
The primary reason for the current limitation is that most DRAs fail to sufficiently leverage the prior knowledge available from public datasets as auxiliary information. Although certain DRA methods like \cite{li2023gan} introduce a Generative Adversarial Network (GAN) pre-trained on the public dataset as an image prior, they merely explore its latent space coarsely and fail to exploit its full potential.

To address this, we propose a novel GAN-based DRA approach, Data Reconstruction Attack via \textbf{P}rogressive \textbf{F}eature \textbf{O}ptimization (PFO), which effectively disassembles the GAN structure and leverages hierarchical features between intermediate blocks. This intuition originates from recent studies \cite{bau2018gan,shen2020interpreting,tewari2020pie,daras2021intermediate} that have demonstrated that there is rich semantic information encoded in the latent code and hierarchical features of GANs. Specifically, we consider the generator of the GAN as a concatenation of multiple basic blocks and the representations produced between the blocks as hierarchical features. To launch the attack, we first perform optimization in the latent space of the generator and then successively optimize the hierarchical features from the first internal block to the last internal block. Furthermore, we introduce an $l_1$ ball constraint to restrict the deviation during optimization, to avoid unreal image generation. During this process, the reconstructed image will progressively approximate the macro-features and the fine-grained features of target images, ultimately achieving pixel-level reconstruction of the privacy-sensitive images. We conduct systematic experiments to evaluate our method at different split points in multiple settings, including high-resolution image recovery, OOD scenarios, and various target models. Moreover, we also conduct assessments against defense strategies and extend our attack to the black-box scenario and heterogeneous data. The encouraging experimental results demonstrate that our proposed method outperforms previous DRA methods on multiple metrics and maintains robustness against the employed defense mechanisms compared to baselines. Our main contributions are summarized as follows:
\begin{itemize}
    \item We propose a novel GAN-based DRA method named PFO, which disassembles the generator into basic blocks and progressively optimizes the latent code and hierarchical features under the $l_1$ ball constraint.
    \item The proposed attack achieves state-of-the-art performance in a range of scenarios, particularly under the high-resolution image recovery, out-of-distribution (OOD) settings, and models with more complex structures.
    \item We conduct extensive experiments to validate the robustness of our method against defense strategies and outstanding transferability on heterogeneous data and black-box scenarios.
\end{itemize}
\section{Background}
\subsection{Split Inference}
Split Inference (SI) evolved from Split Learning (SL) \cite{hauswald2014hybrid,eshratifar2019jointdnn,banitalebi2021auto,kang2017neurosurgeon,matsubara2022split,ko2024two},  aiming to distributively deploy the inference process of structurally complex models across cloud servers and edge devices. This technique simultaneously reduces computational overhead on the endpoint devices and ensures data privacy to some extent, thus exhibiting considerable application potential.

In a typical SI system, a well-trained model $M$ is partitioned into two parts, which are deployed as client model $M_C:\mathcal{X}\xrightarrow{}\mathcal{H}$ for users and cloud model $M_S:\mathcal{H}\xrightarrow{}\mathcal{Y}$ for service providers, respectively. When users input their raw data $\textbf{x}$ into the SI system, the client model $M_C$ first processes it to obtain the intermediate representation $\textbf{h}=M_C(\textbf{x})$. Then,  $\textbf{h}$ is transmitted to the cloud server and mapped into the proper output $y$ through the cloud model $y=M_S(\textbf{h})$.  Finally, the output $y$ is returned to the users to complete the whole inference process. Through collaborative inference between the cloud and the client, the edge-side devices merely execute the client model $M_C$ instead of the full model $M$, facilitating the reduction of computation overhead. Meanwhile, the cloud server only has access to the smashed intermediate representation $\textbf{h}$, providing a certain level of privacy preservation for users. By incorporating the above two advantages, the SI technique is capable of striking a trade-off between privacy and utility.

\subsection{Data Reconstruction Attacks on Split Inference}
Data Reconstruction Attacks (DRAs) enable pixel-level reconstruction of privacy-sensitive features in target images, as shown in Fig.\ref{fig:dra}. Typically, DRAs can be categorized as \textit{Optimization-based} methods and \textit{Learning-based} methods according to attacking mechanisms.

For the \textit{Optimization-based} DRAs, the adversary tries to optimize an initial image $\textbf{x}$ sampled from a random distribution or a pre-trained generator. Since the attacker can only access intermediate representations, the optimization objective for this problem is defined as minimizing the distance between the representations computed from the initial image $\textbf{h}=M_C(\textbf{x})$  and the target representations $\textbf{h}_{tar}$. The optimization problem is formulated as follows:
\begin{equation}
\begin{aligned}
    \textbf{x}^*= \mathop{\arg\min}\limits_{\textbf{x}\in\mathcal{X}} \ d_{\mathcal{H}}(M_C(\textbf{x}), \textbf{h}_{tar})+\lambda R(\textbf{x}),
\label{Eq: optim}
\end{aligned}
\end{equation}
where $d_{\mathcal{H}}(.,.)$ represents the specific metric for measuring the distance between representations, $R(.)$ serves as an auxiliary regularization term to ensure realistic generation of the initial image, and $\lambda$ is a hyperparameter controlling the weight of $R(.)$. Through this optimization process, the final image $\textbf{x}^*$ maximally approximates the target image $\textbf{x}_{tar}$, i.e.,  $\textbf{x}^* \approx \textbf{x}_{tar}$.

The pioneering work, rMLE \cite{he2019model}, first formulated this attack as an optimization problem, minimizing the Euclidean distance between feature representations while using Total Variation (TV) \cite{rudin1992nonlinear} as a prior to ensure the image's visual plausibility. Building on this, Likelihood Maximization (LM) \cite{singh2021disco} significantly improved reconstruction quality by employing a pre-trained autoencoder as a more powerful deep image prior, which constrains the generated image to a more natural manifold.

For the \textit{Learning-based} DRAs, the adversary aims at training an inverse network $M^{-1}: \mathcal{H}\xrightarrow{}\mathcal{X}$ to directly invert $\textbf{h}_{tar}$ back to the image space $\mathcal{X}:\textbf{x}^*=M^{-1}(\textbf{h}_{tar})$. To obtain $M^{-1}$, the attacker tends to utilize a publicly available dataset $D_{pub}$ and process it through the $M_C$ to get a paired dataset $(\textbf{x}_{pub}, M_C(\textbf{x}_{pub}))$. Then, $M^{-1}$ is trained as follows:
\begin{equation}
\begin{aligned}
   M^{-1}_{best}= \mathop{\arg\min}\limits_{M^{-1}} \ d_{\mathcal{X}}(M^{-1}(M_C(\textbf{x}_{pub})), \textbf{x}_{pub})),
\end{aligned}
\end{equation}
 where $d_{\mathcal{X}}(.,.)$ measures the distance between the images. After training, the attacker obtains the final image using the $M^{-1}_{best}$, i.e., $\textbf{x}^*=M^{-1}_{best}(\textbf{h}_{tar})$.
 
A representative method is the Inverse-Network (IN) \cite{he2019model}. This approach involves training a dedicated decoder on feature-image pairs acquired from the client model, which learns a direct mapping that can reconstruct an image from a target representation in a single forward pass.

To counter such threats, various defenses have been developed to protect intermediate representations. Early strategies focused on feature obfuscation; for instance, Siamese Defense \cite{osia2020hybrid} employs a Siamese network to learn inversion-resistant features, while NoPeek \cite{vepakomma2020nopeek} utilizes adversarial training to make representations uninformative to an attacker. Subsequent methods shifted towards direct information reduction, such as DISCO \cite{singh2021disco} which sparsifies feature maps.

\subsection{GAN as prior}
Generative Adversarial Networks (GANs) \cite{goodfellow2014generative} are a class of DNNs composed of a generator and a discriminator. Through adversarial training, they learn the distribution from the training dataset with the objective of generating high-quality and diverse realistic images \cite{goodfellow2020generative}. Currently, prominent examples of GANs include the WGAN \cite{arjovsky2017wasserstein} and StyleGAN series \cite{karras2019style,karras2020analyzing,karras2021alias}. With strong generative capabilities, GANs have been widely used in various privacy attacks \cite{fang2023gifd,qiu2024closer,yu2024calor,fang2024clip,zhuang2025stealthy,yu2025icas}, and were first introduced into the DRA field by GLASS \cite{li2023gan}.

Recent works \cite{bau2018gan,shen2020interpreting,tewari2020pie,daras2021intermediate} have demonstrated the rich semantic information of hierarchical features in GANs, indicating the potential to go deep into the internal structure. Thus, our investigation focuses on exploring the possibility of leveraging the hierarchical space of internal blocks to enhance DRA.
\section{Methodology}
\begin{figure*}[tbp]
\centerline{\includegraphics[width=\linewidth]{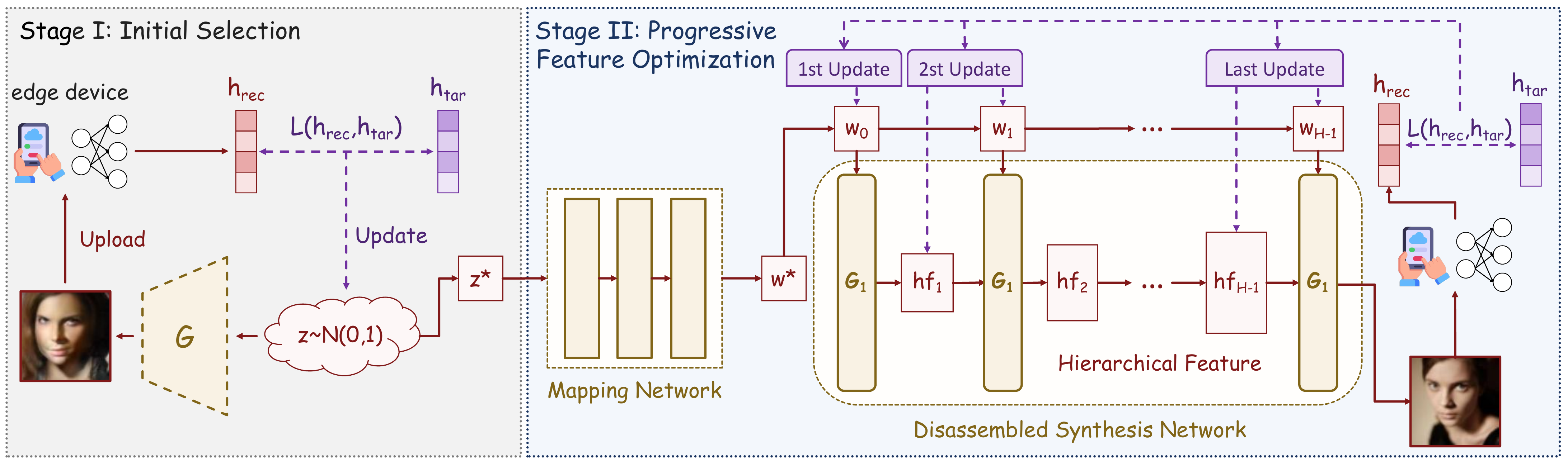}}
\caption{Overview of our proposed PFO method.}
\label{fig:pipeline}
\end{figure*}

In this section, we begin by explaining the threat model of our PFO method in the white-box SI system and introducing the GAN prior employed in our method. Subsequently, we elaborate on the detailed pipeline of the proposed PFO step by step. The overview of our PFO method is shown in Fig. \ref{fig:pipeline}. Finally, we further discuss how to migrate the PFO method to the black-box scenario.

\subsection{Preliminaries}
\textbf{Threat model.} Following previous works \cite{he2019model,singh2021disco,yang2022measuring,li2023gan,xu2024stealthy}, we assume the cloud service provider acts as an honest-but-curious member in the SI system, whose goal is to reconstruct the target private image $\textbf{x}_{tar}$ from the received intermediate representations $\textbf{h}_{tar}$. Considering that service providers tend to possess greater privileges in real-world scenarios, we account for the worst-case setting where the attacker can gain white-box access to the full parameters and architecture of the target model $M$ (including the client part $M_C$ and the server part $M_S$). Besides, the Internet hosts a vast number of open-source datasets, where the adversary can readily retrieve essential datasets $D_{pub}$ with similar distribution to the target dataset $D_{tar}$ (e.g., selecting other facial datasets as auxiliary datasets when attacking a facial recognition dataset) and use them to train generative priors. In this paper, our proposed PFO is an optimization-based DRA method and utilizes a GAN prior to provide high-fidelity images. Therefore, the overall optimization process of PFO follows Eq. (\ref{Eq: optim}) and the optimized image $\textbf{x}$ is produced by the GAN prior $\textbf{x}=G(\textbf{z})$, where $\textbf{z}$ is the input latent vector of the generator.

\textbf{GAN prior.} Our PFO method relies on the pre-trained StyleGAN2 \cite{karras2020analyzing} as a generative prior. The generator $G$ of StyleGAN2 consists of two components, which are a mapping network $G_{map}:\mathcal{Z}\xrightarrow{}\mathcal{W}$ and a synthesis network $G_{syn}:\mathcal{W}\xrightarrow{}\mathcal{X}$, respectively. The mapping network $G_{map}$ maps the initial latent vectors $\mathbf{z}\in{\mathcal{Z}}$ into the extended $\mathcal{W}$ space \cite{abdal2019image2stylegan}, while the synthesis network $G_{syn}$ generates images $\mathbf{x}$ using the extended vectors $\mathbf{w}\in{\mathcal{W}}$. According to the previous research \cite{abdal2019image2stylegan},  the reduced feature entanglement in $\mathcal{W}$ space facilitates better style generation. Thus, we also optimize $\mathbf{w}$ during the progressive feature optimization. Notably, the StyleGAN2 is pre-trained without the utilization of the target model $M$ or other auxiliary prior corresponding to the target dataset, making it more flexible when attacking different target models and datasets under OOD scenarios.

\subsection{Data Reconstruction Attack via Progressive Feature Optimization}
\textbf{Initial Selection.}  Considering the randomness in sampling latent vectors $\mathbf{z}$, it is obvious that not all of them facilitate the generation of appropriate images. However, for the optimization of non-convex functions, a proper initial point is extremely significant to determine the whole optimization direction. To mitigate the risk of generating low-quality and misleading images, it is critical to perform initial selection in $\mathcal{Z}$ space and $\mathcal{W}$ space to avoid local minima and obtain reliable latent vectors. Specifically, we first randomly sample a large number of $\mathbf{z}$ as candidates and then optimize them as follows:
\begin{equation}
\begin{aligned}
    \textbf{z}^* = \mathop{\arg\min}\limits_{\mathbf{z}\in\mathcal{Z}} \ L_{match}(M_C(\textbf{x}_{tar}), M_C(G(\textbf{z}))) \\+\lambda R(\textbf{z})+\alpha \mathcal{TV}(G(\textbf{z})),
    \label{full loss}
\end{aligned}
\end{equation}
where $L_{match}(.,.)$ is the Mean Square Error (MSE) loss between the target representation and the representation computed from the generated image $G(\textbf{z})$, serving as the main guidance for $\textbf{z}$. $\lambda$ and $\alpha$ are hyperparameters to determine the weight of regularization terms, respectively. To avoid unreal generation when using MSE loss, we introduce KL-based regularization \cite{kingma2013auto} to constrain the $\textbf{z}$ to ensure it conforms to the standard Gaussian distribution:
\begin{equation}
\begin{aligned}
    R(\textbf{z}) = -\frac{1}{2} \sum_{i=1}^k \left(1 + \log(\sigma_i^2) - \mu_i^2 - \sigma_i^2\right),
\end{aligned}
\end{equation}

where $\mu_i^2$ and $\sigma_i^2$ are the element-wise mean and standard deviation of the $\textbf{z}$. Additionally, we adopt Total Variation \cite{rudin1992nonlinear} to encourage piece-wise smooth generation as follows:
\begin{equation}
\begin{aligned}
    \mathcal{TV}(\textbf{x})=\sum_{i,j} \sqrt{|\textbf{x}_{i+1,j} - \textbf{x}_{i,j}|^2 + |\textbf{x}_{i,j+1} - \textbf{x}_{i,j}|^2},
\end{aligned}
\end{equation}
where $\textbf{x}=G(\textbf{z})$. By combining the above loss terms, it can be ensured that the final $\textbf{z}^*$ is a relatively ideal starting point to generate the initial extended vector $\textbf{w}_{init}=G_{map}(\textbf{z}^*)$ for the following stage. 

\textbf{Progressive Feature Optimization.} Previous research \cite{karras2020analyzing} proves that the front blocks in the StyleGAN2 control the overall characteristics, while the back ones concentrate on local details. Therefore, we can leverage the internal blocks of the generator to achieve progressive optimization. The complete workflow of progressive feature optimization is displayed in Algorithm \ref{pseu}. 

First, we optimize $\textbf{w}_{init}$ to obtain the optimal start point $\textbf{w}_0$. The synthesis network $G_{syn}$ is split into $H+1$  internal blocks, i.e., $G_{syn}=G_{H+1}\circ{G}_{H}\circ\cdots{G}_{2}\circ{G}_{1}$.  Each block contains two styled convolution modules \cite{karras2019style}, which are the basic generation units in the StyleGAN2. Then, we may input the $\textbf{w}_0$ into the first block $G_1$ to obtain the first hierarchical feature $\mathbf{hf}_{1}^0$.  For each internal block $G_{i+1},i\in[1,\dots,H]$, its hierarchical feature $\mathbf{hf}_{i+1}^{0}$ is computed by following steps:
\begin{itemize}
    \item Initially, we generate images utilizing the remaining blocks ($i.e., \mathbf{x}_{i}^{j-1}={G}_{remain}(\mathbf{hf}_{i}^{j-1},\mathbf{w}_{i}^{j-1})$) and feed them into $M_C$ to compute the loss with the target intermediate representation $\textbf{h}_{tar}$.
    \item Then, we repeat the former step to iteratively update both $\mathbf{w}_{i}$ and $\mathbf{hf}_{i}$. Meanwhile, we restrict the $\mathbf{hf}_{i}$ within the $l_1$ ball with radius  ${r}[{i}]$ centered at the initial hierarchical feature $\mathbf{hf}_{i}^{0}$, which acts as a correction device to avoid exceeding shift that may lead to collapse generation.
    \item Once the iteration is finished, the final $\mathbf{w}^{N}_{i}$ and $\mathbf{hf}^{N}_{i}$ will be fed into the next block $G_{i+1}$ for the next hierarchical feature $\mathbf{hf}_{i+1}^{0}$. Notaly, we denote the $\mathbf{w}^{N}_{i}$ as the initial vector $\mathbf{w}_{i+1}^{0}$ before the next optimization starts.
\end{itemize}
After all the hierarchical features are optimized, the final image $\textbf{x}^*$ can be obtained by $\textbf{x}^*=\textbf{hf}_{H+1}^0=G_{H+1}(\textbf{hf}_{H}^N)$.

\begin{algorithm}[!ht]
\caption{Pseudocode of progressive feature optimization}\label{pseu}
\begin{algorithmic}[1]
\renewcommand{\algorithmicrequire}{\textbf{Input:}}
\renewcommand{\algorithmicensure}{\textbf{Output:}}
\REQUIRE $G_{syn}$: the synthesis network; $H$: the number of hierarchical features; $M_C$: the target client model; $\textbf{h}_{tar}$: the target intermediate representation; $\mathcal{L}$: the loss function; $r[1\dots{L}]$: the radius value of $l_1$ ball for each hierarchical features; $N$: the number of iterations;
\ENSURE Reconstructed images $\mathbf{x}^*$;
\STATE Acquire extended vector $\mathbf{w}_{init}$ via initial selection
\STATE $\mathbf{w}_{0} \xleftarrow{} \mathop{\arg\min}\limits_{\mathbf{w}} \mathcal{L}(M_C(G_{syn}(\mathbf{w}_{init})), \textbf{h}_{tar})$
\STATE Disassemble the $G_{syn}$ into $G_{H+1}\circ{G}_{H}\circ\cdots{G}_{2}\circ{G}_{1}$
\STATE Obtain the first hierarchical feature $\mathbf{hf}_{1}^0=G_{1}(\mathbf{w}_{0})$
\STATE Set $\mathbf{w}_{1}^{0}=\mathbf{w}_{0}$
\FOR{$i\xleftarrow{}1 \ $to$ \ H$} 
    \STATE Set $G_{remain}=G_{H+1}\circ G_{H}\dots \circ G_{i+1}$
    \FOR{$j\xleftarrow{}1 \ $to$ \ N$}
        \STATE $loss = \mathcal{L}(M_C(G_{remain} (\mathbf{hf}_{i}^{j-1},\mathbf{w}_{i}^{j-1})), \textbf{h}_{tar})$
        \STATE $\mathbf{hf}_{i}^{j} \xleftarrow{} Optimizer(\mathbf{hf}_{i}^{j-1};loss), ||\mathbf{hf}_{i}^{j}-\mathbf{hf}_{i}^0||_1 \leq r[i]$
        \STATE $\mathbf{w}_{i}^{j} \xleftarrow{} Optimizer(\mathbf{w}_{i}^{j-1};loss), ||\mathbf{w}_{i}^{j}-\mathbf{w}_{i}^0||_1 \leq r[i]$
    \ENDFOR
    \STATE $\mathbf{hf}_{i+1}^{0} = G_{i+1}(\mathbf{hf}_{i}^N,\mathbf{w}_{i}^N)$
    \STATE $\mathbf{w}_{i+1}^{0} = \mathbf{w}_{i}^N$
\ENDFOR
\STATE The final image $\mathbf{x}^* = \mathbf{hf}_{H+1}^{0}$ 
\RETURN $\mathbf{x}^*$
\end{algorithmic}
\end{algorithm}

Since an $l_1$-ball constraint is employed during the progressive feature optimization to constrain the distribution of extended vectors $\textbf{w}$ and hierarchical feature $\textbf{hf}$, we remove the KL divergence term from the loss function in this stage. Hence,  the loss function $\mathcal{L}$ is formulated as follows:
\begin{equation}
\begin{aligned}
    \mathcal{L} = L_{match}(\textbf{h}_{tar},\textbf{h}_{rec}) +\alpha \mathcal{TV}(G_{remain}(\textbf{hf},\textbf{w})),
\end{aligned}
\end{equation}
where $\textbf{h}_{rec}=M_C(G_{remain}(\textbf{hf},\textbf{w}))$.

\subsection{Extension to Blackbox Scenario}
In the black-box setting, the malicious attacker loses access to the model's parameters and architectures, retaining only the ability to query it for intermediate representations.  At this point, the adversary is unable to compute gradients to perform backpropagation, thus leading to the urgent need for seeking gradient-free optimization methods. 

For our proposed PFO method, we select the typical Covariance Matrix Adaptation (CMA) \cite{hansen2016cma} optimizer as an alternative to gradient-based optimizers. This optimizer employs gradient-free search to estimate the direction for parameter optimization, thus allowing our approach to execute under black-box settings. The corresponding experiment in Sec. \ref{sec: blackbox} shows that our method maintains the best performance compared to baselines under the black-box scenario.

\begin{table*}[htbp]
  \setlength{\tabcolsep}{1pt}
  \normalsize
  \centering
  \caption{Overall comparison of our method with state-of-the-art methods against ResNet-18 trained on CelebA.}
  \begin{threeparttable} 

    \resizebox{\linewidth}{!}{
    \begin{tabular}{lcccccccccccc} \toprule
    \multicolumn{1}{c}{\multirow{2}[0]{*}{Method}} & \multicolumn{4}{c}{Split Point 1} & \multicolumn{4}{c}{Split Point 2} & \multicolumn{4}{c}{Split Point 3}\\
    \cmidrule{2-5}\cmidrule(lr){6-9}\cmidrule(lr){10-13} & PSNR$\uparrow{}$ & MSE$\downarrow{}$ & SSIM$\uparrow{}$ & LPIPS$\downarrow{}$ & PSNR$\uparrow{}$ & MSE$\downarrow{}$ & SSIM$\uparrow{}$ & LPIPS$\downarrow{}$ & PSNR$\uparrow{}$ & MSE$\downarrow{}$ & SSIM$\uparrow{}$ & LPIPS$\downarrow{}$\\ \midrule
    rMLE & 24.084 & 0.023 & 0.731 & 0.266 & 21.294 & 0.044 & 0.603 & 0.413 & 14.101 & 0.242 & 0.310 & 0.642 \\
    LM & 28.937 & 0.010 & 0.886 & 0.118 & 24.024 & 0.028 & 0.732 & 0.281 & 19.119 & 0.089 & 0.532 & 0.511 \\
    IN & 25.128 & 0.021 & 0.764 & 0.241 & 22.950 & 0.034 & 0.694 & 0.279 & 20.755 & 0.059 & 0.642 & 0.292 \\
    GLASS & 29.340 & 0.007 & 0.885 & 0.090 & 25.480 & 0.017 & 0.801 & 0.146 & 20.320 & 0.065 & 0.636 & 0.239 \\
    \textbf{PFO(ours)} & \textbf{33.103} & \textbf{0.003} & \textbf{0.945} & \textbf{0.048} & \textbf{27.426} & \textbf{0.012} & \textbf{0.857} & \textbf{0.118} & \textbf{21.803} & \textbf{0.042} & \textbf{0.697} & \textbf{0.223} \\
    \bottomrule
    \end{tabular}}
    \end{threeparttable} 
  \label{table:main}%
\end{table*}%
\section{Experiments}
In this section, we conduct comprehensive experiments to validate the effectiveness of our proposed method. We first illustrate the details of our experimental settings. Then, we make an overall comparison between our method and state-of-the-art baselines to evaluate the attack performance. Furthermore, we conduct extensive experiments on more challenging scenarios to assess the generalizability of our method in various settings. Finally, we perform an ablation study to assess the contributions of each component in our method.

\subsection{Experimental Setup}
\textbf{Models.} We evaluate our attack against a diverse set of target models ($M_C$), including the ResNet-18, ResNet-152 \cite{he2016deep} and the CLIP-RN50 vision-language model \cite{radford2021learning}. These models are assumed to be privately trained and deployed by a service provider. For the image reconstruction, we employ a pre-trained StyleGAN2 \cite{karras2020analyzing} as a strong generative prior, and we utilize StyleGAN-XL \cite{sauer2022stylegan} in heterogeneous scenarios.

\textbf{Datasets.} We utilize (1) CelebA \cite{liu2015deep}, which contains 202,599 face images of 10,177 identities (2) FFHQ \cite{karras2019style}, which consists of 70,000 high-quality face images with considerable variation in terms of age, ethnicity, and image background. We conduct facial reconstruction experiments at two scales: for standard-resolution ($64 \times 64$) analysis, the downstream task is binary attractiveness classification on CelebA; for high-resolution ($224 \times 224$) analysis, the task is 1000-class classification. Furthermore, to evaluate performance on heterogeneous data, we employ (3) CIFAR-10 \cite{krizhevsky2009learning} and (4) CINIC-10 \cite{darlow2018cinic}, which are established 10-class object recognition datasets.

\begin{figure}[tbp]
\centerline{\includegraphics[width=\linewidth]{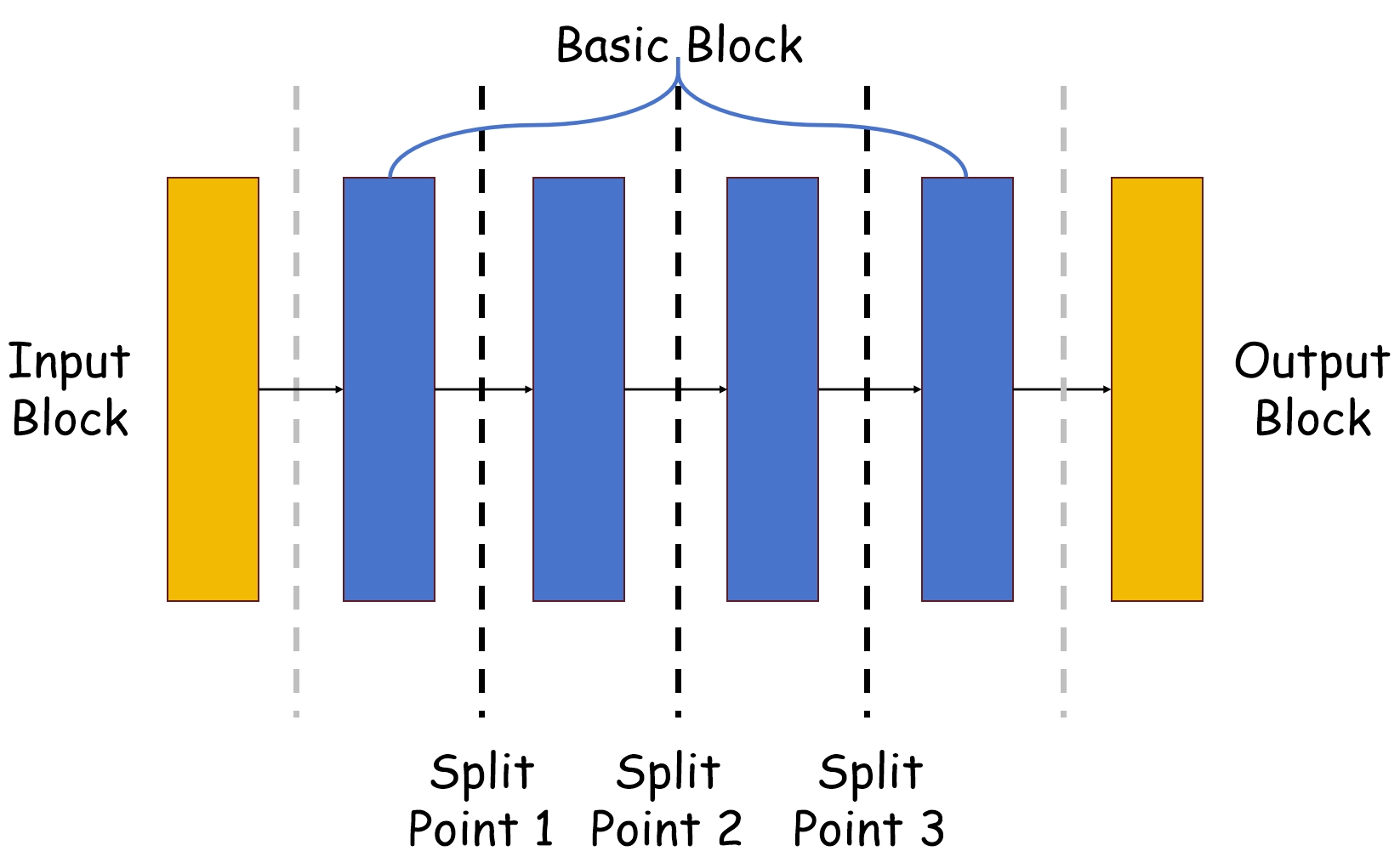}}
\caption{The split points in our experiments.}
\label{fig:split}
\end{figure}

\textbf{Split Point.} As shown in Fig. \ref{fig:split}, we divide the target model into six blocks and present three split points. The blocks before the split point are treated as the client model $M_C$ and the blocks after the split point serve as the cloud model $M_S$, respectively.

\textbf{Compared Baselines \& Defenses.} We benchmark our method against four prominent data reconstruction attacks (DRAs): regularized Maximum Likelihood Estimation (rMLE) \cite{he2019model}, Likelihood Maximization (LM) \cite{singh2021disco}, Inverse-Network (IN) \cite{he2019model}, GLASS \cite{li2023gan}. To further assess the robustness of our approach, we evaluate its performance against four representative privacy-preserving defenses, including Siamese Defense \cite{osia2020hybrid}, NoPeek \cite{vepakomma2020nopeek}, DISCO \cite{singh2021disco}, and Noise Mask \cite{titcombe2021practical}.

\textbf{Attack Setup.} Our experimental setup involves distinct private and public datasets. For experiments on CelebA, we partition it into a private set $D_P$ (80,525 images from 3,000 identities) for training the target model, and a public set $D_A$ (the remaining images) for training the StyleGAN2. This ensures no identity overlap between the datasets. To evaluate transferability, we also conduct a cross-dataset attack where the StyleGAN2 is trained on the public FFHQ dataset and used to attack the model trained on CelebA.

We configure the hyperparameters for our attacks and the baselines to ensure a fair comparison. For optimization-based methods, including our own, the learning rate is set to $1 \times 10^{-2}$ for 20,000 iterations. The TV regularization for rMLE and LM is set to 2.0 and 1.5, respectively, while for our GAN-based attack it is set to 0.01. For the learning-based IN, the inversion model is trained for 30 epochs with a learning rate of $1 \times 10^{-3}$. 

When evaluating against defenses, we select a challenging hyperparameter setting for each mechanism. For DISCO, we use a privacy-utility trade-off $\rho=0.95$ and a channel pruning ratio of $R=0.1$. For Noise Mask, we inject Laplacian noise with location $a=0$ and scale $b=1.0$. The trade-off parameter for NoPeek is set to $\lambda_2=5$, and for Siamese Defense, the weight of the Siamese loss is set to $\lambda_3=0.005$.

\textbf{Evaluation Metrics} We evaluate the similarity between the original and reconstructed images using four complementary metrics to comprehensively quantify the reconstruction performance. We employ Mean Squared Error (MSE) and Peak Signal-to-Noise Ratio (PSNR) \cite{hore2010image} to measure pixel-wise fidelity. To better align with human perceptual judgment, which is crucial for assessing privacy risks, we also adopt the Structural Similarity Index (SSIM) \cite{wang2004image} and the Learned Perceptual Image Patch Similarity (LPIPS) \cite{zhang2018unreasonable}. For MSE and LPIPS, lower values indicate a more successful reconstruction, whereas for PSNR and SSIM, higher values are better.

\begin{table}[!ht]
    \setlength{\tabcolsep}{5pt}
    \normalsize
    \centering
    \caption{Comparison results against ResNet-18 models trained on CelebA with the public dataset being FFHQ, i.e., the GAN prior is pre-trained on the FFHQ dataset.}
    \label{table:ffhq64-celeba64-resnet18}
    \begin{threeparttable} 
    \resizebox{\linewidth}{!}{
    \begin{tabular}{cccccc}
        \toprule
         \textbf{Split Point} & \textbf{Method} & PSNR$\uparrow{}$ & MSE$\downarrow{}$ & SSIM$\uparrow{}$ & LPIPS$\downarrow{}$ \\ \midrule
         \multirow{2}{*}{Split Point 1} 
         & GLASS & 27.89 & 0.010 & 0.848 & 0.123 \\
         & \textbf{PFO(ours)} & \textbf{31.862} & \textbf{0.005} & \textbf{0.922} & \textbf{0.066} \\
         \midrule 
         \multirow{2}{*}{Split Point 2}
         & GLASS & 23.39 & 0.027 & 0.701 & 0.210 \\
         & \textbf{PFO(ours)} & \textbf{24.883} & \textbf{0.020} & \textbf{0.769} & \textbf{0.186} \\
         \midrule 
         \multirow{2}{*}{Split Point 3}
         & GLASS & 17.85 & 0.108 & 0.453 & 0.336\\
         & \textbf{PFO(ours)} & \textbf{18.862} & \textbf{0.081} & \textbf{0.537} & \textbf{0.318}\\
         \bottomrule
    \end{tabular}
    }
    \end{threeparttable}
\end{table}
\begin{table}[!ht]
    \setlength{\tabcolsep}{5pt}
    \normalsize
    \centering
    \caption{Comparison results against the ResNet-152 model trained on CelebA with the public dataset being FFHQ. In this table, the resolution of images from CelebA and FFHQ is 224×224 and 256×256 respectively, serving as the high-resolution setting.}
    \label{table:ffhq256-celeba224-resnet152}
    \begin{threeparttable} 
    \resizebox{\linewidth}{!}{
    \begin{tabular}{cccccc}
        \toprule
         \textbf{Split Point} & \textbf{Method} & PSNR$\uparrow{}$ & MSE$\downarrow{}$ & SSIM$\uparrow{}$ & LPIPS$\downarrow{}$ \\ \midrule
         \multirow{2}{*}{Split Point 1} 
         & GLASS & 19.361 & 0.005 & 0.589 & 0.364 \\
         & \textbf{PFO(ours)} & \textbf{21.647} & \textbf{0.003} & \textbf{0.711} & \textbf{0.240} \\
         \midrule 
         \multirow{2}{*}{Split Point 2}
         & GLASS & 17.039 & 0.010 & 0.550 & 0.412 \\
         & \textbf{PFO(ours)} & \textbf{19.496} & \textbf{0.005} & \textbf{0.658} & \textbf{0.311} \\
         \midrule 
         \multirow{2}{*}{Split Point 3}
         & GLASS & 15.823 & 0.013 & 0.494 & 0.459\\
         & \textbf{PFO(ours)} & \textbf{16.023} & \textbf{0.012} & \textbf{0.560} & \textbf{0.415}\\

         \bottomrule
    \end{tabular}
    }
    \end{threeparttable}
\end{table}

\subsection{Overall Comparison on Attack Performance}
We compare our PFO method with the previous state-of-the-art methods, which are rMLE \cite{he2019model}, LM \cite{singh2021disco}, IN \cite{he2019model}, and GLASS \cite{li2023gan}. All the hyperparameters of the compared baselines are aligned with the recommended values in their official papers. 

\begin{figure}[tbp]
\centerline{\includegraphics[width=\linewidth]{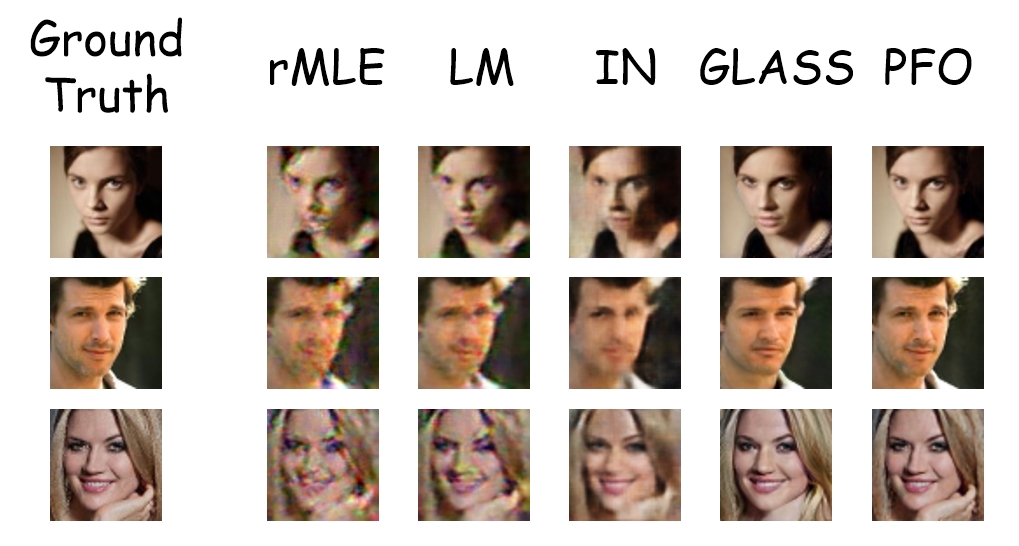}}
\caption{Visualization of different DRA methods under the split point 2.}
\label{fig:vis}
\end{figure}

The overall comparison results are listed in the Table \ref{table:main}. The target model is ResNet-18 trained on the private set of CelebA and the generative prior is pre-trained on the public set of CelebA. The resolution of images from the public dataset and private dataset is set to 64×64, following the common settings in the previous works \cite{he2019model,singh2021disco,yang2022measuring,li2023gan}. From the quantitative results in Table \ref{table:main}, we can observe that our PFO method outperforms all the baselines on all four metrics under the three split points. Taking the PSNR metric as an example, our method achieves an improvement of at least 1.5 dB and up to 4 dB over the previous state-of-the-art methods, highlighting the superiority of the PFO method.

Visualization of the reconstructed images is displayed in Fig. \ref{fig:vis}. Our reconstructed images demonstrate higher fidelity and realism thwan baselines, further validating the superiority of exploiting GAN’s intermediate features.

\begin{table}[!ht]
    \setlength{\tabcolsep}{5pt}
    \normalsize
    \centering
    \caption{Comparison results against the ResNet-152 model trained on FaceScrub with the public dataset being FFHQ. As the high-resolution setting, the resolution of images from FaceScrub and FFHQ is 224×224 and 256×256, respectively.}
    \label{table:ffhq256-facescrub224-resnet152}
    \begin{threeparttable} 
    \resizebox{\linewidth}{!}{
    \begin{tabular}{cccccc}
        \toprule
         \textbf{Split Point} & \textbf{Method} & PSNR$\uparrow{}$ & MSE$\downarrow{}$ & SSIM$\uparrow{}$ & LPIPS$\downarrow{}$ \\ \midrule
         \multirow{2}{*}{Split Point 1} 
         & GLASS & 20.003 & 0.005 & 0.665 & 0.406 \\
         & \textbf{PFO(ours)} & \textbf{23.012} & \textbf{0.003} & \textbf{0.795} & \textbf{0.265} \\
         \midrule 
         \multirow{2}{*}{Split Point 2}
         & GLASS & 16.087 & 0.013 & 0.610 & 0.485 \\
         & \textbf{PFO(ours)} & \textbf{19.423} & \textbf{0.006} & \textbf{0.747} & \textbf{0.337} \\
         \midrule 
         \multirow{2}{*}{Split Point 3}
         & GLASS & 15.069 & 0.015 & 0.545 & 0.516\\
         & \textbf{PFO(ours)} & \textbf{15.568} & \textbf{0.013} & \textbf{0.631} & \textbf{0.434}\\

         \bottomrule
    \end{tabular}
    }
    \end{threeparttable}
\end{table}
\begin{table}[!ht]
    \setlength{\tabcolsep}{5pt}
    \normalsize
    \centering
    \caption{Comparison results against the CLIP-RN50 model. Both of the target model and the GAN prior are trained on the FFHQ dataset (private and public parts), with the resolution set to 256×256.}
    \label{table:ffhq256-resnet50}
    \begin{threeparttable} 
    \resizebox{\linewidth}{!}{
    \begin{tabular}{cccccc}
        \toprule
         \textbf{Split Point} & \textbf{Method} & PSNR$\uparrow{}$ & MSE$\downarrow{}$ & SSIM$\uparrow{}$ & LPIPS$\downarrow{}$ \\ \midrule
         \multirow{2}{*}{Split Point 1} 
         & GLASS & 15.227 & 0.015 & 0.650 & 0.368 \\
         & \textbf{PFO(ours)} & \textbf{16.014} & \textbf{0.013} & \textbf{0.825} & \textbf{0.150} \\
         \midrule 
         \multirow{2}{*}{Split Point 2}
         & GLASS & 14.010 & 0.020 & 0.600 & 0.408 \\
         & \textbf{PFO(ours)} & \textbf{15.136} & \textbf{0.016} & \textbf{0.786} & \textbf{0.170} \\
         \midrule 
         \multirow{2}{*}{Split Point 3}
         & GLASS & 14.466 & 0.018 & 0.584 & 0.384\\
         & \textbf{PFO(ours)} & \textbf{16.582} & \textbf{0.012} & \textbf{0.743} & \textbf{0.168}\\
         \bottomrule
    \end{tabular}
    }
    \end{threeparttable}
\end{table}

\subsection{Evaluation in Challenging Scenarios}
In this part, we further investigate the effectiveness of our method in more challenging scenarios. Due to the page limit, subsequent tables will only present experimental results for our PFO method and GLASS. The reason why we select GLASS as the baseline is that it is the state-of-the-art method among prior approaches as shown in the Table \ref{table:main}.\\

\textbf{Out-of-Distribution (OOD) Settings.} Unlike the common setting \cite{li2023gan} where the public dataset shares a similar distribution with the private dataset (e.g., dividing the CelebA dataset into disjoint parts), the OOD setting indicates that the two datasets have large distribution shift with each other. Specifically, the two datasets are totally different datasets instead of two slices from a single large dataset. The corresponding results are shown in Table \ref{table:ffhq64-celeba64-resnet18}, \ref{table:ffhq256-celeba224-resnet152} and \ref{table:ffhq256-facescrub224-resnet152}. 

Table \ref{table:ffhq64-celeba64-resnet18} shows the evaluation results of  low-resolution (64×64) OOD settings. Compared to Table \ref{table:main}, our PFO method and GLASS both suffer a decrease in the overall performance, while PFO method still maintains the best performance. Table \ref{table:ffhq256-celeba224-resnet152} and \ref{table:ffhq256-facescrub224-resnet152} display the evaluation results of high-resolution (224×224) OOD settings. It is obvious that our PFO presents better transferability than the baseline in both high-resolution and low-resolution scenarios.

\textbf{High-resolution Image Recovery Settings.} Previous works \cite{he2019model,singh2021disco,yang2022measuring,li2023gan} merely concentrate on the easier low-resolution (64×64) scenarios. Therefore, we increase the resolution to 224×224 to better suit real-world applications.  The experimental results for high-resolution evaluation are listed in Table \ref{table:ffhq256-celeba224-resnet152}, \ref{table:ffhq256-facescrub224-resnet152}, and \ref{table:ffhq256-resnet50}.

Table \ref{table:ffhq256-celeba224-resnet152} and \ref{table:ffhq256-facescrub224-resnet152} validate the high-resolution image recovery under the harder OOD setting, while Table \ref{table:ffhq256-resnet50} assesses the performance under the non-OOD setting. The results presented in the three tables demonstrate that our method achieves the best under the high-resolution settings.,

\begin{table}[!ht]
    \setlength{\tabcolsep}{5pt}
    \normalsize
    \centering
    \caption{Comparison results against the ResNet-18 model trained on the CelebA dataset under the black-box setting.}
    \label{table:blackbox}
    \begin{threeparttable} 
    \resizebox{\linewidth}{!}{
    \begin{tabular}{cccccc}
        \toprule
         \textbf{Split Point} & \textbf{Method} & PSNR$\uparrow{}$ & MSE$\downarrow{}$ & SSIM$\uparrow{}$ & LPIPS$\downarrow{}$ \\ \midrule
         \multirow{2}{*}{Split Point 1} 
         & GLASS & 18.366 & 0.090 & 0.564 & 0.268 \\
         & \textbf{PFO(ours)} & \textbf{19.263} & \textbf{0.072} & \textbf{0.591} & \textbf{0.261} \\
         \midrule 
         \multirow{2}{*}{Split Point 2}
         & GLASS & 16.667 & 0.146 & 0.491 & 0.312 \\
         & \textbf{PFO(ours)} & \textbf{17.502} & \textbf{0.115} & \textbf{0.534} & \textbf{0.287} \\
         \midrule 
         \multirow{2}{*}{Split Point 3}
         & GLASS & 13.378 & 0.302 & 0.346 & 0.371\\
         & \textbf{PFO(ours)} & \textbf{14.316} & \textbf{0.260} & \textbf{0.393} & \textbf{0.357}\\

         \bottomrule
    \end{tabular}
    }
    \end{threeparttable}
\end{table}
\begin{table}[!ht]
    \setlength{\tabcolsep}{5pt}
    \normalsize
    \centering
    \caption{Comparison results against the ResNet-18 model trained on CINIC-10.The StyleGAN-XL prior is trained on CIFAR-10.}
    \label{table:hetegeneous}
    \begin{threeparttable} 
    \resizebox{\linewidth}{!}{
    \begin{tabular}{cccccc}
        \toprule
         \textbf{Split Point} & \textbf{Method} & PSNR$\uparrow{}$ & MSE$\downarrow{}$ & SSIM$\uparrow{}$ & LPIPS$\downarrow{}$ \\ \midrule
         \multirow{2}{*}{Split Point 1} 
         & GLASS & 15.249 & 0.706 & 0.258 & 0.369 \\
         & \textbf{PFO(ours)} & \textbf{16.801} & \textbf{0.488} & \textbf{0.362} & \textbf{0.292} \\
         \midrule 
         \multirow{2}{*}{Split Point 2}
         & GLASS & 14.436 & 0.857 & 0.223 & 0.375 \\
         & \textbf{PFO(ours)} & \textbf{14.937} & \textbf{0.767} & \textbf{0.256} & \textbf{0.327} \\
         \midrule 
         \multirow{2}{*}{Split Point 3}
         & GLASS & 13.683 & 1.083 & 0.154 & 0.435\\
         & \textbf{PFO(ours)} & \textbf{13.972} & \textbf{0.976} & \textbf{0.162} & \textbf{0.412}\\
         \bottomrule
    \end{tabular}
    }
    \end{threeparttable}
\end{table}

\textbf{Assessment of More Complex DNNs.} Except for the commonly used ResNet-18 model in previous works \cite{he2019model,singh2021disco,yang2022measuring,li2023gan}, we further employ complex models for evaluation, including ResNet-152 and CLIP-RN50. Related experimental results are shown in Table \ref{table:ffhq256-celeba224-resnet152}, \ref{table:ffhq256-facescrub224-resnet152}, and \ref{table:ffhq256-resnet50}.

According to the results in the aforementioned three tables, our PFO method is excellent in adapting to the complex models. However, we notice that the CLIP-RN50 model causes the most reduction in the performance even in the non-OOD scenario. The potential reason for this phenomenon might be the different tasks where the model is trained, e.g., the CLIP-RN50 is trained as a vision foundation for multiple vision tasks, while the ResNet-152 is trained for simple classification tasks. This highlight indicates the promising future work on how the tasks affect the performance of DRAs.

\subsection{Extended Evaluation}
Apart from the above evaluation, we conduct more experiments to comprehensively validate the generalizability of our PFO method. By default, the target model is the ResNet-18 trained on the private set of CelebA and the GAn prior is pre-trained on the public set of CelebA, which are the same as the main experiment of Table \ref{table:main}.

\begin{table}[!ht]
    \setlength{\tabcolsep}{5pt}
    \normalsize
    \centering
    \caption{Comparison results against the ResNet-18 model trained on CelebA with defenses. The experiments in this table are performed on Split Point 1.}
    \label{table:defense}
    \begin{threeparttable} 
    \resizebox{\linewidth}{!}{
    \begin{tabular}{cccccc}
        \toprule
         \textbf{Defense Strategy} & \textbf{Method} & PSNR$\uparrow{}$ & MSE$\downarrow{}$ & SSIM$\uparrow{}$ & LPIPS$\downarrow{}$ \\ \midrule
         \multirow{2}{*}{No Peek} 
         & GLASS & 21.160 & 0.091 & 0.701 & 0.233 \\
         & \textbf{PFO(ours)} & \textbf{22.441} & \textbf{0.052} & \textbf{0.785} & \textbf{0.195} \\
         \midrule 
         \multirow{2}{*}{DISCO}
         & GLASS & 28.905 & 0.008 & 0.879 & 0.099 \\
         & \textbf{PFO(ours)} & \textbf{31.236} & \textbf{0.005} & \textbf{0.922} & \textbf{0.068} \\
         \midrule 
         \multirow{2}{*}{Siamese}
         & GLASS & 28.847 & 0.011 & 0.889 & 0.108 \\
         & \textbf{PFO(ours)} & \textbf{29.060} & \textbf{0.009} & \textbf{0.929} & \textbf{0.079} \\
         \bottomrule
    \end{tabular}
    }
    \end{threeparttable}
\end{table}
\begin{table}[!ht]
    \setlength{\tabcolsep}{5pt}
    \normalsize
    \centering
    \caption{Ablation study performed on ResNet-18 trained on the CelebA dataset. "w/o" is the abbreviation for "without".}
    \label{table:ablation}
    \begin{threeparttable} 
    \resizebox{\linewidth}{!}{
    \begin{tabular}{cccccc}
        \toprule
         \textbf{Split Point} & \textbf{Method} & PSNR$\uparrow{}$ & MSE$\downarrow{}$ & SSIM$\uparrow{}$ & LPIPS$\downarrow{}$ \\ \midrule
         \multirow{3}{*}{Split Point 1} 
         & w/o PFO & 28.292 & 0.009 & 0.867 & 0.106 \\
         & PFO w/o $l_1$-ball & 32.184 & 0.005 & 0.918 & 0.075 \\
         & \textbf{PFO} & \textbf{33.103} & \textbf{0.003} & \textbf{0.945} & \textbf{0.048} \\
         \midrule 
         \multirow{3}{*}{Split Point 2}
         & w/o PFO & 24.523 & 0.022 & 0.767 & 0.168 \\
         & PFO w/o $l_1$-ball & 24.449 & 0.022 & 0.786 & 0.207 \\
         & \textbf{PFO} & \textbf{27.426} & \textbf{0.012} & \textbf{0.857} & \textbf{0.118} \\
         \midrule 
         \multirow{3}{*}{Split Point 3}
         & w/o PFO & 18.180 & 0.113 & 0.549 & 0.301 \\
         & PFO w/o $l_1$-ball & 18.714 & 0.083 & 0.587 & 0.348 \\
         & \textbf{PFO} & \textbf{21.803} & \textbf{0.042} & \textbf{0.697} & \textbf{0.223} \\

         \bottomrule
    \end{tabular}
    }
    \end{threeparttable}
\end{table}

\textbf{Defense Analysis.} Following prior works \cite{li2023gan}, we select three typical defenses against DRAs to analyze the robustness of our proposed PFO, which are NoPeek \cite{vepakomma2020nopeek}, Siamese Defense \cite{osia2020hybrid} and DISCO \cite{singh2021disco}. The results are listed in the Table \ref{table:defense}. Due to the limited paper length, we only present the evaluation results in the Split Point 1.

As shown in Table \ref{table:defense}, our PFO method maintains the superior robustness against the three defenses, compared to the baseline. Notably, the DISCO and Siamese defenses are weak in resisting the current state-of-the-art DRAs, while the NoPeek still maintains effectiveness to some extent. This result indicates that with the continuous improvement in DRA performance, stronger defenses accordingly need to be continuously studied and updated.

\label{sec: blackbox}
\textbf{Black-box Setting.} Under the black-box scenario, the adversary has no access to the parameters and detailed architectures of the target model, thereby unable to compute the gradients for backpropagation. We substitute the gradient-based optimizer with gradient-free optimizer CMA \cite{hansen2016cma} to further evaluate whether the proposed method can be effective under the black-box scenario.

Table \ref{sec: blackbox} demonstrates the evaluation results of black-box attacks. Although both methods suffer a significant decrease in the attack performance, our method still outperforms the baseline method GLASS. This result reveals the challenges of black-box scenarios, shedding light on future research directions.

\textbf{Heterogeneous Data.} The above experiments are all conducted on the facial datasets. Following the prior work \cite{li2023gan}, we extend our evaluation to the heterogeneous data, such as natural images in CINIC-10 dataset. The results are listed in the Table \ref{table:hetegeneous}. Our PFO demonstrates better performance than the baseline method.

\begin{figure}[tbp]
\centerline{\includegraphics[width=\linewidth]{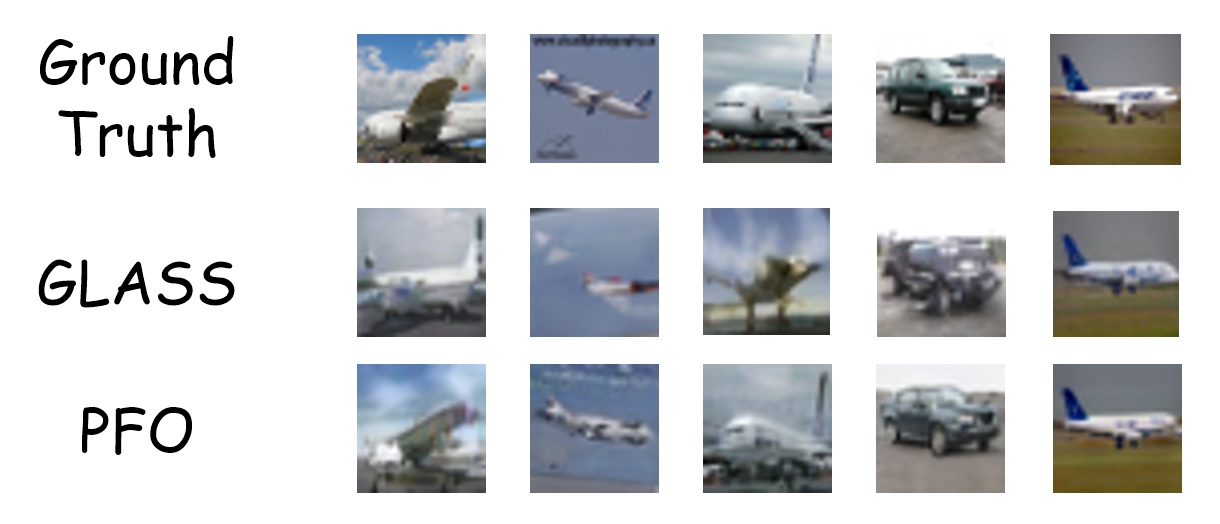}}
\caption{Visualization of heterogeneous data.}
\label{fig:het}
\end{figure}

Visualization of heterogeneous data is shown in Fig .\ref{fig:het}.

\subsection{Ablation Study}
To estimate the contributions from each component in our method, we conduct ablation studies on the ResNet-18 trained on the private set of CelebA dataset using the StyleGAN2 trained on the public dataset of CelebA. The overall results are listed in the Table \ref{table:ablation}. 

\textbf{Progressive Feature Optimization.} We remove the progressive feature optimization stage from our pipeline while keeping the remaining paradigm unchanged. The corresponding setting is named "w/o PFO". From the results between "w/o PFO" and PFO, we can observe that there is a significant degradation up to 4 dB in PSNR metric. Other metrics have also shown varying degrees of decline. Therefore, utilizing the hierarchical features plays a critical role in our pipeline.

\textbf{$l_1$-ball constraint.} To avoid unreal image generation, we introduce the $l_1$-ball constraint into the progressive features optimization. The setting of removing $l_1$-ball is named "PFO w/o $l_1$-ball". As shown in Table \ref{table:ablation}, the $l_1$-ball is beneficial in improving the performance in all metrics. Thus, we demonstrate the necessity of restricting the hierarchical features and $\textbf{w}$ vectors within the $l_1$-ball space.
\section{Conclusion and Future Work}
In this paper, we introduce PFO, a novel and strong data reconstruction attack designed for split inference systems. Through extensive experiments, we demonstrate that PFO consistently outperforms existing methods. Its efficacy has been validated across a wide spectrum of challenging scenarios, including against prominent defenses like NoPeek and DISCO, in challenging high-resolution settings, and even in cross-model and black-box configurations. These findings reveal profound privacy vulnerabilities in modern deep learning models deployed in split inference frameworks and highlight that current defense mechanisms may offer insufficient protection against powerful generative attacks.

While PFO sets a new benchmark for reconstruction fidelity, we identify its computational overhead—a common trait of optimization-based methods—as a key limitation. This presents clear avenues for future research. An important direction may involve enhancing the attack's efficiency, perhaps by developing hybrid models that integrate our optimization strategy with faster learning-based techniques. Furthermore, our findings highlight the urgent need for novel defense mechanisms specifically designed to counteract attacks that exploit powerful generative priors. Additionally, extending the evaluation of PFO to other emerging model architectures, such as Vision Transformers, will be crucial for comprehensively understanding and mitigating the evolving privacy risks in next-generation split inference systems.

\textbf{Societal Impact and Ethical Considerations.} A potential negative impact could be malicious users leveraging the proposed attack method to reconstruct private data from the public system. To alleviate potential dilemma, a cautious approach is to establish access permissions.

\vspace{12pt}

\end{document}